\documentclass[11pt,a4paper]{article}
\usepackage{times,latexsym}
\usepackage[T1]{fontenc}
\usepackage[utf8]{inputenc}
\usepackage{url}

\usepackage{amsmath}
\usepackage{amssymb,enumitem}
\usepackage{mathtools}
\usepackage{amsfonts}
\usepackage{amsthm}
\usepackage{graphicx}
\usepackage{booktabs}
\usepackage{tabularx}
\usepackage{makecell}
\usepackage{enumitem}
\usepackage{nicefrac}
\usepackage{microtype}
\usepackage{xcolor}
\usepackage[most]{tcolorbox}
\usepackage{float}
\usepackage{caption}
\usepackage{textcomp}
\usepackage{newunicodechar}
\usepackage{multirow}
\usepackage{listings}
\usepackage{placeins} 
\newunicodechar{–}{--}
\newunicodechar{—}{---}
\newunicodechar{‘}{`}
\newunicodechar{’}{'}
\newunicodechar{“}{``}
\newunicodechar{”}{''}

\usepackage[acceptedWithA]{tacl2021v1}

\usepackage[capitalize,noabbrev]{cleveref}

\usepackage{etoolbox}
\apptocmd{\thebibliography}{\interlinepenalty=10000\relax\microtypesetup{protrusion=false}}{}{\PackageWarning{main}{could not patch thebibliography}}

\hypersetup{pdfauthor={Ashutosh Srivastava, Siddharth Yedlapati, Vinay Aggarwal, Yaman Kumar Singla, Shashwat Dixit, Jitendra Ajmera, Balaji Krishnamurthy},
  pdftitle={Benchmarking the Personalization Capabilities of Large Language Models},
  pdfkeywords={}}

\definecolor{boxTitleBg}{HTML}{8E3A36}
\definecolor{boxBodyBg}{HTML}{F6D9D6}
\definecolor{green}{HTML}{0B6623}

\theoremstyle{plain}

\theoremstyle{definition}

\theoremstyle{remark}

\lstdefinestyle{promptbox}{
  basicstyle=\footnotesize\ttfamily,
  breaklines=true,
  breakatwhitespace=false,
  breakindent=1.2em,
  postbreak=\mbox{\textcolor{boxTitleBg}{$\hookrightarrow$}\space},
  columns=fullflexible,
  keepspaces=true,
  showstringspaces=false,
  aboveskip=3pt, belowskip=1pt, xleftmargin=2pt,
}

\usepackage{fancyhdr}
\fancypagestyle{titlepage}{%
  \fancyhf{}%
  \fancyfoot[L]{\footnotesize\url{https://huggingface.co/datasets/behavior-in-the-wild/SDR-Bench}}%
}

\title{Benchmarking the Personalization Capabilities of Large Language Models}

\author{
Ashutosh Srivastava \quad Siddharth Yedlapati \quad Vinay Aggarwal \quad Yaman Kumar Singla \\
\bfseries Shashwat Dixit \quad Jitendra Ajmera \quad Balaji Krishnamurthy \\[0.3em]
\normalfont\raisebox{-0.20em}{\includegraphics[height=1.4em]{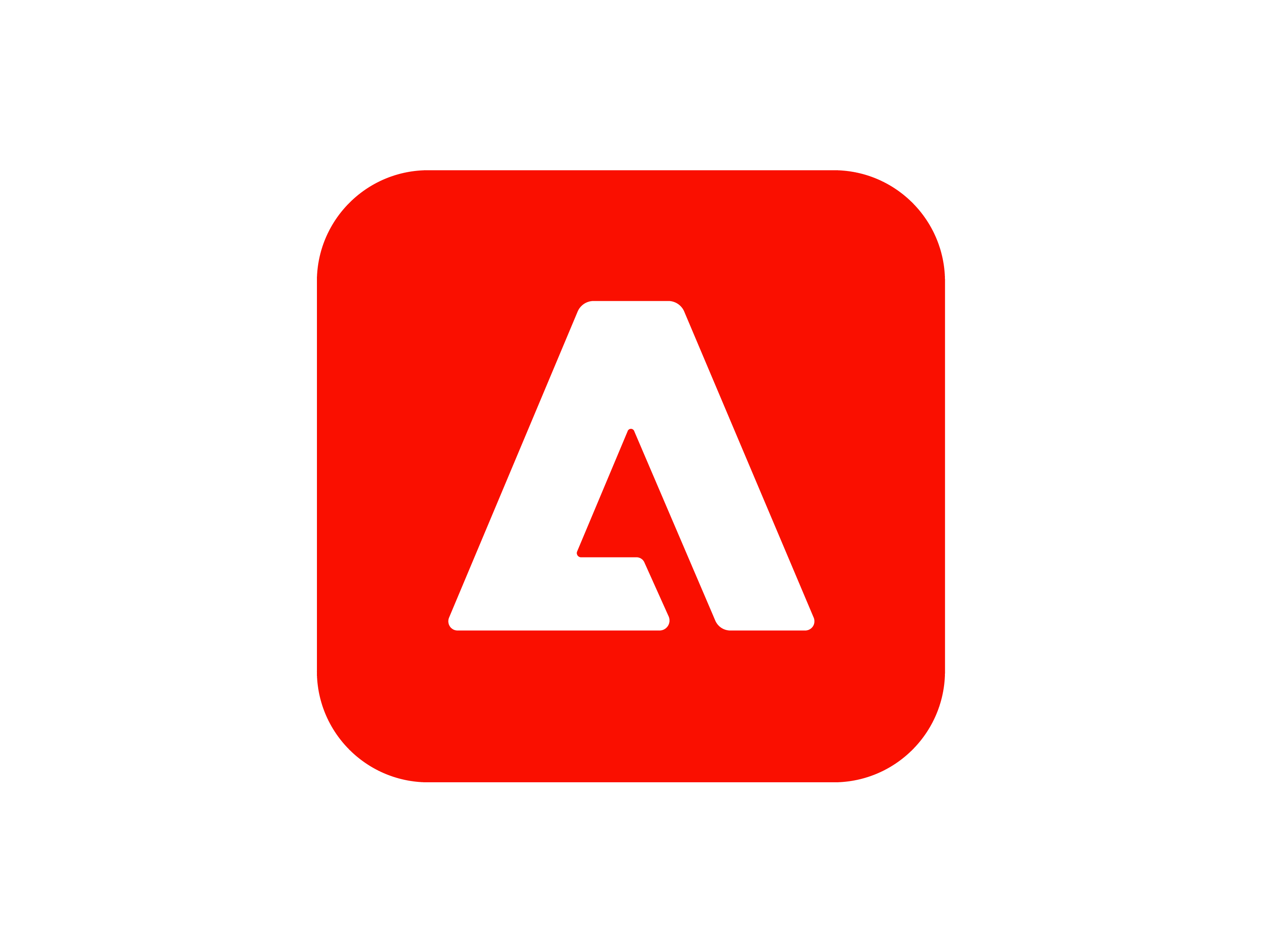}}\hspace{0.15em}Adobe Media \& Data Science Research \\
\normalfont\texttt{behavior-in-the-wild@googlegroups.com}
}

\begin{document}

\maketitle
\thispagestyle{titlepage}

\begin{abstract}

\looseness=-1 Personalization is classically a two-party problem: a sender chooses what to say, and a receiver with independent objectives decides whether to act. A salesperson pitching the same analytics product leads with HIPAA compliance for a hospital and real-time reporting for a retailer, expecting a different argument to work on each. Existing LLM personalization benchmarks measure a narrower, one-party property: whether output matches the preferences of the same user it serves—sender and receiver being the same, as when RLHF aligns an assistant to its own user. The two-party case is harder to study automatically, since it needs ground truth linking specific content to an observed receiver action.

\looseness=-1 Sales outreach provides this: a message written for one prospect, recorded against whether it produced a reply, a call, or a closed deal. We introduce SDR-Arena, a framework for benchmarking two-party generative personalization at scale, and SDR-Bench, a public corpus of 50,000 customer success stories across 22 industries and 3500 enterprises. Given only pre-outcome information, an agent must reconstruct the arguments that won the deal, scored by a weighted nugget-recall metric (WCS). The best model, Claude Sonnet 4.6, reaches 55.8\% WCS indicating it recovers only half the winning content—a plateau we observe across model families that costly deep-research pipelines do not close. An ablation shows the cause is retrieval, not reasoning: models improve substantially given the facts a human researcher would gather, but rarely find them through web search alone. Two studies with professional SDRs support the metric: only 48\% of generated pitches were rated usable without editing, and WCS yields model rankings consistent with evaluation against expert strategies authored independently of any model output.

\end{abstract}

\begin{figure*}[t]
    \centering
    \includegraphics[width=\textwidth]{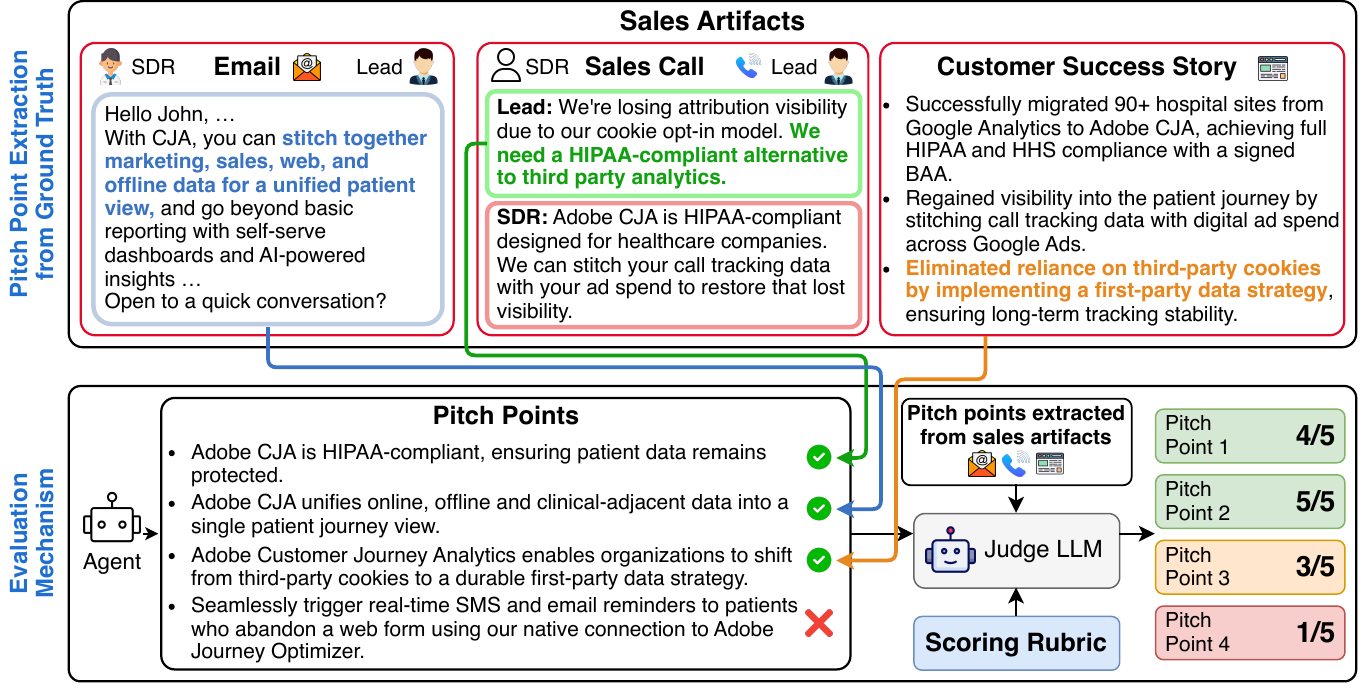}
    \caption{Overview of SDR-Arena showcasing how LLM generated output is compared with artifacts like Sales Emails, Transcripts \& Success Stories to benchmark their personalization capability}
    \label{fig:e2e-overview}
\end{figure*}
\section{Introduction}

\label{sec:intro}

\looseness=-1 Communication as defined by the seminal work of \citet{lasswell1948structure} characterizes any communicative act as \emph{who says what to whom, through which channel, with what effect, and at what time}. Within this framework, personalization can be defined as the act of varying the message, that is, the \textit{what}, while conditioning on (and holding fixed) the speaker, receiver, channel, and time. The act of communication has two parties whose interests need not initially align; a sender, who selects the message with the goal of inducing some action from the receiver, and a receiver, who has independent preferences and chooses whether to act. The study of how senders shape messages to induce receiver action has been a long tradition across psychology, economics, and marketing. The Yale Communication and Attitude Change program \citep{hovland1953communication} established that persuasion depends jointly on the source, the message, and the audience. The Elaboration Likelihood Model \citep{PETTY1986123} later distinguished persuasion driven by argument quality from the one driven by surface cues such as tone or delivery, a distinction we adopt directly in \Cref{sec:problem_formulation}. Bayesian Persuasion \citep{kamenica2011bayesian} formalized the sender-receiver relationship itself as a signaling game, in which a sender chooses what to reveal in order to shift a receiver's beliefs and, in turn, their actions. 

\looseness=-1 Machine learning research on personalization, by contrast, has largely sidestepped this two-party dynamic, approaching personalization instead as learning a policy to retrieve and rank items from a fixed inventory of candidates for a single user. Recommender systems rank items from a catalog against user-history signals \citep{ricci2010introduction}; advertising platforms optimize the selection, and targeting of ads among a fixed pool of pre-authored creatives \citep{choi2020onlinedisplay}; and persona-based dialogue systems condition responses on explicit persona representations within narrow domains \citep{zhang-etal-2018-personalizing}.

Large language models transition personalization from a retrieval and ranking problem to a generative one: rather than selecting from a fixed inventory, an LLM can generate a continuum of message variants for a given (speaker, receiver, channel, time) tuple, conditioned on whatever receiver attributes can be inferred from context. A growing body of applied work has begun to exploit this shift. LLM-generated messages tailored to an inferred psychological profile shift attitudes and purchase intent more than generic messaging \citep{matz2024potential}. LLM-personalized learning materials improve student
motivation and performance \citep{tasdelen2025generative}, and assistants increasingly
adapt to individual users \citep{chen2024large}. These results raise a natural
question: how well do current LLMs personalize in the classical sense---as a
sender selecting what to say to induce a specific receiver's action?

\textbf{Related Work.}
Existing evaluations of LLM personalization, however, largely measure a different, one-party property: how well a model's output aligns with the stated preferences of the \emph{same} user it serves, rather than whether that output changes the behavior of an independent receiver. LaMP \citep{salemi2024lamp} evaluates personalized text generation conditioned on a user's history of past interactions. PersonaConvBench \citep{li2025personalized} scores persona-grounded conversational quality. PersonaLens \citep{zhao2025personalens} evaluates assistant behavior under declared user preferences. \citet{li2025personalized, zhao2025personalens, salemi2024lamp, afzoon2024persobench, jiang2025personamem} all share this structure: the recipient of the generated message is the user issuing the prompt, and the optimization target is alignment
between output and that user's preferences, in the same sense that RLHF aligns an assistant to its own user. 

The two-party question, whether generated content changes the behavior of a receiver with independent objectives, has instead been studied almost exclusively through randomized human-subject experiments, in which participants are exposed to human- or LLM-generated persuasive messages and evaluated on subsequent shifts in attitudes, agreement, or behavioral intentions \citep{matz2024potential, durmus2024persuasion}. Such studies are tied to a specific methodology, take weeks to run, cannot be re-run against a new model on demand, and yield comparisons specific to individual studies rather than portable across systems. There is thus a need for rigorous, automated, and scalable methods to study two-party generative
personalization.

This requires an empirical setting where receiver actions are observed and logged against the specific messages that induced them, messages are authored at the receiver level, and ground truth linking generated content to receiver action is available at scale. 
Sales outreach offers exactly this: outreach is drafted one-to-one by a Sales Development Representative (SDR) for a specific prospect, and receiver actions (replying, scheduling a call, closing a deal) are routinely logged against the outreach that induced them \citep{tehro2022}. Personalization here is concrete: an SDR selling the same analytics product might emphasize regulatory compliance to a healthcare prospect and real-time reporting to a retailer. We call each such argument---a specific claim connecting the product capability to a prospect's need, a \textit{pitch point}, and an outreach a set of such pitch points chosen to induce the receiver's action. The sales funnel also produces human-authored artifacts at each successful transition (emails that secured a call,
call transcripts that opened deal discussions, and post-deal success stories),
each serving as ground truth for successful personalization at a different stage of the same (seller, prospect, product) tuple, with failed outreach providing a natural negative class against which any proposed metric can be validated.

We introduce \textbf{SDR-Arena} (Fig.~\ref{fig:e2e-overview}), the first framework, to our knowledge, to benchmark two-party generative personalization at scale, instantiated in sales to evaluate whether an LLM agent, standing in for the SDR, can recover the pitch points that historically induced a successful outcome, using only the information available before that outcome was known. Our contributions are:

\begin{itemize}[leftmargin=*, nosep]
    \item \textbf{SDR-Arena: A benchmark for two-party generative personalization}. We operationalize the task as recovering the pitch points that historically induced a successful outreach outcome, and score agent output against them using an LLM-judged coverage metric, giving the field an automated, repeatable substitute for the weeks-long human-subject studies two-party personalization has previously required.
    
    \item \textbf{SDR-Bench}, a public corpus of 50,000 customer success stories from  3500 enterprises across 22 industries, each paired with the seller, prospect, product, and historical timestamp required to evaluate whether an agent can recover the right information and generate the strategic content of the deal-closing pitch, enabling reproducible, cross-model comparison. 

    \item \textbf{An enterprise-scale evaluation,} applying SDR-Arena to proprietary sales email and call-transcript corpora from a Fortune-100 tech company and a mid-sized healthcare firm (115,000 filtered outreach emails and 5,435 calls by 127 SDRs, labeled by outcome), testing whether our findings on public success stories hold on unfiltered, live-pipeline outreach, rather than only on post-hoc, publicly published success stories.

    \item \textbf{A personalization plateau.} Benchmarking frontier LLMs
    and deep-research agents on SDR-Bench and on real enterprise outreach
    shows that even the best-performing model, Claude Sonnet~4.6, recovers
    only about $50\%$ of the winning strategic content, a gap that persists across model families and is not closed by costly deep-research pipelines. A controlled ablation traces the cause to query formulation rather than reasoning: models improve substantially when handed the facts an expert researcher would gather, but fail to surface them through web search on their own.
    
    \item \textbf{Expert Validation} via field deployment with 12 professional SDRs and a gold-standard exercise with senior SDRs from five enterprises who author reference strategies independently of any model output, showing our metric tracks real expert judgment (Pearson $r=0.82$).
\end{itemize}

We release the benchmark to foster further research on generative personalization.

\begin{figure*}[t]
    \centering
    \includegraphics[width=\textwidth]{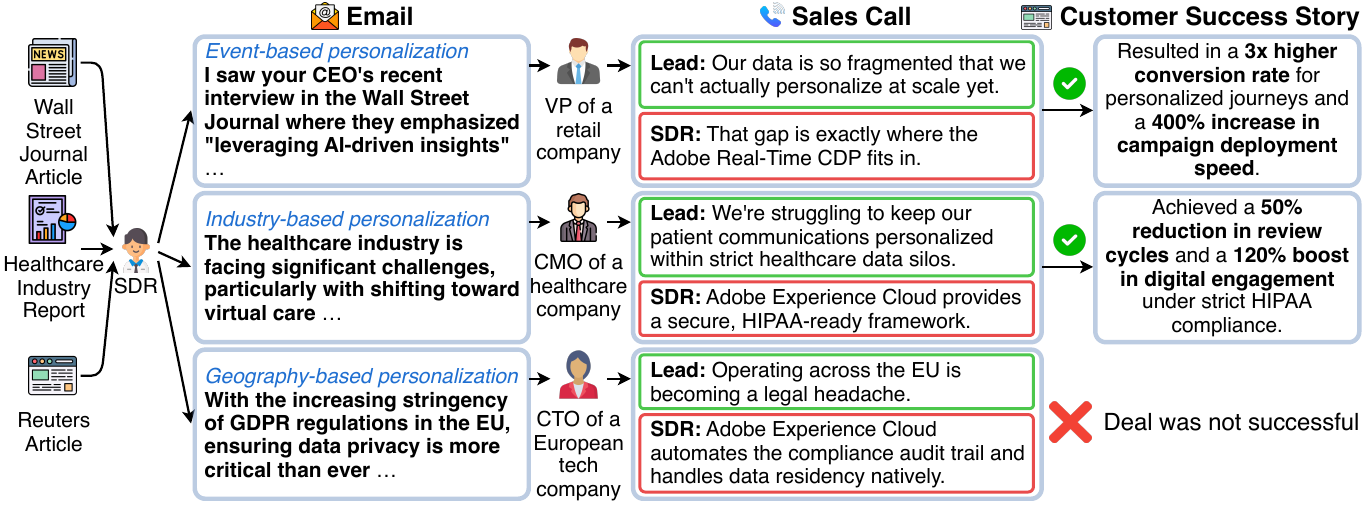}
    \caption{Sales Journey: From Prospecting to Outreach to Call and eventual Deal Closure leading to Success Story publication}
    \vspace{-10pt}
    \label{fig:Sales Journey}
\end{figure*}

\section{Problem Formulation}
\label{sec:problem_formulation}

We now instantiate the two-party formulation of \Cref{sec:intro} in sales, turning it into a concrete task with observable ground truth. We first describe the sales development lifecycle that produces this ground truth (\Cref{sec:lifecycle}); we then define the agent's prediction task (\Cref{sec:eval-methodology}), the content-coverage metric used to score it (\Cref{sec:eval_framework}), and the temporally constrained environment for back-testing the agents (\Cref{sec:sdr_arena}).

\subsection{Sales Development Lifecycle}
\label{sec:lifecycle}

A sales journey (Fig.~\ref{fig:Sales Journey}) begins with SDRs researching prospective accounts for needs and budget signals, then sending personalized outreach emails to induce the prospect to agree to a sales call; an outreach email is considered successful if it leads to one. The call surfaces the prospect's needs in more depth and progresses toward deal closure, while unsuccessful outreach receives no response.

Following a successful closure, the workflow often culminates in a \textbf{Customer Success Story}: a publicly documented case study published by the seller company showcasing how their products helped a customer overcome key challenges. Released as web articles, these stories validate the partnership by documenting the transition from a `pain state' to a `success state' \citep{tehro2022}. Examples include success stories from \citet{oracle-careem}, \citet{salesforce-snapology}, and \citet{adobe-marriott}.

\looseness=-1 Each stage of this funnel leaves a human-authored artifact at a \emph{known successful transition}: an email that secured a call, a transcript that opened deal discussions, and a success story that documents a closed deal. Because every artifact is logged against the specific (seller, prospect, product) interaction that produced it, the funnel yields ground truth for what content actually induced a receiver to act, at scale.

\subsection{The Prediction Task}\label{sec:eval-methodology}

We cast two-party personalization as a historical \emph{back-testing} problem: Can an agent, standing in for the SDR, reconstruct the arguments that historically induced a successful outcome, using only information available before that outcome was known?

Formally, an agent $\Phi$ plays the role of the sender. Each evaluation instance is a tuple $(S, C, P, t)$: a seller $S$, a prospective customer $C$, the product(s) $P$ being pitched, and the historical timestamp $t$ of the interaction. Given this tuple and a view of the public web $W_t$ restricted to information available at time $t$ (\Cref{sec:sdr_arena}), the agent researches the prospect and produces an outreach represented as a set of \emph{pitch points}:
\begin{equation}\label{eq:policy}
\hat{O} = \{\hat{pp}_1, \dots, \hat{pp}_n\} = \Phi(S, C, P, W_t).
\end{equation}
\looseness=-1 A pitch point, as introduced in Section 1, is an atomic argument that links a product capability to a specific need or pain point of the prospect through a concrete value proposition, for example, ``the product is HIPAA-compliant, so patient data stays protected'' connects a compliance capability to a regulatory pain point.

\looseness=-1 Crucially, the agent operates under the same information constraint as the original SDR. it sees only what would have been available at time $t$; it does \emph{not} observe the outreach that was actually sent, $O^*$, nor the recipient's subsequent response. Its task is to generate, from context alone, arguments that are most likely to move this prospect. We then score $\hat{O}$, the generated pitch points, against the pitch points of the historical outreach that \emph{induced} the action, so performance measures how much of that winning argument content the agent recovers.

\subsection{Measuring Personalization through Content Coverage}\label{sec:eval_framework}

\looseness=-1 \textbf{Content, not style.} Effective personalization in enterprise sales is fundamentally about \emph{what} is argued, the reasons connecting a seller's solution to a buyer's situation rather than \emph{how} the message is styled. We ground this in the Elaboration Likelihood Model (ELM) of persuasion \citep{PETTY1986123}, which distinguishes a \emph{central route}, where persuasion under high elaboration is driven by argument quality, from a \emph{peripheral route}, where persuasion under low elaboration is driven by surface cues such as tone and formatting. Enterprise B2B
purchasing is a canonical high-elaboration setting, where high-stakes,
expert buyers scrutinize argument content over long cycles, so whether the
buyer acts is governed primarily by content, with stylistic delivery
secondary. Measuring the argument content an agent produces is therefore a
valid basis for evaluating personalization in this domain.

\looseness=-1\textbf{Nugget-based coverage.} To measure how well an agent's content
covers the arguments that matter, we adopt nugget-based evaluation, which
decomposes a reference text into atomic concepts (``nuggets'') and scores a
candidate by the fraction it covers, following the Pyramid method
\citep{nenkova2004pyramid} and the TREC QA evaluations \citep{voorhees2003trecqa}, and
more recently adapted to LLM-judged nugget recall for RAG evaluation
\citep{pradeep2025nugget}. ELM identifies argument content as the object of
interest; nugget recall measures the fraction of that content a generated
outreach conveys relative to a historically successful one.

\textbf{Ground-truth nuggets.} For every logged success, we take the human-authored outreach $O^*$ that induced the action and extract its pitch points $V^* = \{pp_1, \dots, pp_N\}$. These content nuggets \emph{empirically} moved a real receiver to act, and recovering them is the evaluation objective. We extract these with GPT-4o
\citep{openai2024gpt4o}, prompting an ontological extraction in which each
nugget follows a strict triad structure (Product/Service $\to$ Specific
Pain Point $\to$ Value Proposition/Mechanism) and is validated with an
exact evidence quote from the source text (precision 0.92, recall 0.97
against human annotation; details in Section~\ref{sec:human-study-pitch-points}).

\textbf{Coverage Judge.} We score an agent by weighted nugget recall over $V^*$. An LLM \emph{Coverage Judge} grades how well the agent's output $\hat{O}$ covers each ground-truth nugget $pp_i$ on a six-level scale $s_i \in \{0,\dots,5\}$: \emph{0 (Miss / Irrelevant)}, \emph{1 (Marketing Fluff)}, \emph{2 (Topic Match)}, \emph{3 (Implied / Soft Match)}, \emph{4 (Strong Sales Argument)}, and \emph{5 (Strategic Bullseye)}. A score of $5$ requires that the agent capture both the prospect's specific pain point and the concrete mechanism by which the product addresses it. The full rubric and judge prompt are provided in the Appendix.

\textbf{Weighted Coverage Score.} The per-nugget grades are aggregated into the \textbf{Weighted Coverage Score (WCS)}, the normalized coverage of $V^*$:
\begin{equation}\label{eq:wcs}
\text{WCS} = \left( \frac{\sum_{i=1}^{N} s_i}{5N} \right) \times 100\%.
\end{equation}
WCS measures the agent's \emph{completeness} in recovering the winning sales logic. Because each ground-truth artifact post-dates $t$, this is a strict back-test: a WCS of $100\%$ would mean the agent reconstructed every deal-winning argument using only information available before the outcome was known. An agent that achieves high WCS has thus demonstrated that it can recover the specific strategic content of a message known to have succeeded.

\textbf{WCS is a lower bound on personalization ability.} Since WCS credits only content that matches the historically recorded winner $V^*$, it can \emph{underestimate} an agent that discovers a different strategy that would also have won the deal: the counterfactual outcome of an unsent message is unobservable. Therefore, WCS can be considered a lower bound on the generative personalization ability of LLMs in the two-party setting. One concern is that this undercrediting hides genuine personalization competence: perhaps models produce valid winning arguments that simply differ from $V^*$. In \Cref{sec:discriminative} we test this directly with an outcome-based discriminative task whose scoring does not depend on matching $V^*$ at all; even there, models struggle to identify winning outreach, indicating that the low coverage reflects a real limitation in personalization capability rather than an artifact of measuring coverage against ground truth.

\subsection{SDR-Arena: The Evaluation Environment}
\label{sec:sdr_arena}

We run agents inside \textbf{SDR-Arena}, an evaluation environment to benchmark agents on the public SDR-Bench corpus and the proprietary enterprise data (details in \Cref{sec:dataset}) while eliminating the central threat to any back-testing benchmark: leakage of the very outcome the agent is asked to predict.

\looseness=-1\textbf{Historical Internet Simulator.} If an agent could simply retrieve the success story it is asked to predict, the task would be trivial. To prevent this, every run is served through a Historical Internet Simulator ($W_t$): an isolated environment that enforces a strict temporal boundary on the public web via the \texttt{search\_start\_date} and \texttt{search\_end\_date} parameters to the \citet{brightdata_serp_api} SERP API. Restricting results to time $t$ ensures that an agent's pitch points uses only context available to a human researcher at the time of the original interaction, and never on the not-yet-published record of the deal's success.

\textbf{Agent configurations.} Within this environment we evaluate three configurations, chosen to separate what a model \emph{knows} from what it can \emph{find}:
\begin{itemize}[leftmargin=*,nosep]
    \item \textbf{Closed book LLMs:} models without tools, testing how much strategic content they produce from parametric knowledge alone.
    \item \textbf{LLMs $+$ Web Search:} the same models with tool-augmented search over the simulator, measuring the marginal value of retrieval over parametric generation.
    \item \textbf{Deep Research Agents:} specialized systems that plan, search iteratively, and synthesize retrieved evidence into a comprehensive report.
\end{itemize}
Each configuration receives the same $(S,C,P,t)$ tuple and is prompted to act as a sales representative for seller $S$ pitching product $P$ to prospect $C$.

\section{Dataset Construction}
\label{sec:dataset}

To evaluate generative personalization in real-world settings, we construct two complementary datasets: (i) \textbf{SDR-Bench}, a large-scale public benchmark derived from enterprise sales artifacts, and (ii) a \textbf{private enterprise dataset} containing real sales outreach emails and downstream sales outcomes from two organizations. Together, these datasets provide both reproducible public evaluation and high-fidelity validation on real-world personalized communication.

\textbf{SDR-Bench Dataset.}
\looseness=-1 To enable reproducible public benchmarking of generative personalization systems, we construct SDR-Bench, a large-scale corpus of publicly available customer success stories and testimonials published by enterprises. A success story is a public article documenting how the seller enterprise and its products helped a customer enterprise. We targeted 3500 global enterprises with revenues exceeding \$1B, collecting 7 million webpages with roughly 400 candidate success story URLs. 

\begin{figure*}[t]
    \centering
    \begin{minipage}[t]{0.39\textwidth}
        \centering
        \includegraphics[width=\linewidth]{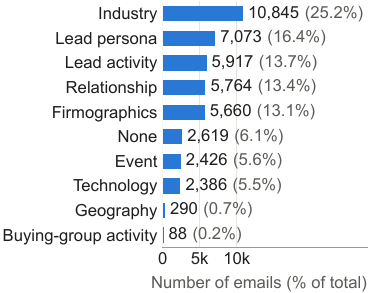}
        \captionof{figure}{Distribution of count of strategies across a random subset of 34,000 emails}
        \label{fig:email_strategy}
    \end{minipage}\hfill
    \begin{minipage}[t]{0.59\textwidth}
        \centering
        \includegraphics[width=\linewidth]{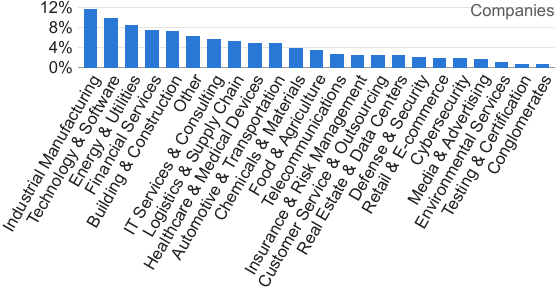}
        \captionof{figure}{SDR-Bench: Distribution of count of companies by Industry Type}
        \label{fig:sdrbench_company_dist_by_industry}
    \end{minipage}
\end{figure*}

A multi-stage filtering pipeline (Table~\ref{tab:sdr_bench_filter}) removed non-text formats, generic landing pages, articles lacking verifiable publication dates or identifiable product solutions, and anonymized stories where the customer organization was not explicitly named (e.g., ``a large food products company''). After URL and locale-based de-duplication, this process yielded a final corpus of 50,000 success-story articles. Distributions of companies and success stories by industry are shown in Fig.~\ref{fig:sdrbench_company_dist_by_industry} and Appendix Fig.~\ref{fig:sdrbench_story_dist_by_industry}. Detailed construction steps are provided in Appendix~\ref{sec:sdr-bench-details}.

\textbf{Private Enterprise Dataset.}
\looseness=-1 To validate that our evaluation metric tracks real world success, we require settings where the ground-truth message $O^*_i$ and its successful outcome $(a=1)$ are explicitly observed. We therefore collaborated with two enterprises---a Fortune 100 technology company and a mid-sized healthcare firm---to collect real human-authored sales outreach paired with downstream prospect actions.

\looseness=-1 For the Fortune 100 company, we filtered down to approximately 90k outreach emails authored by 124 SDRs and 5,435 sales call transcripts over a two-year period (2023--2025), identifying 13,236 instances in which the outreach successfully induced a sales call. For the healthcare firm, we filtered down to 24,506 outreach emails, of which 354 resulted in a scheduled sales call. These successful outreach instances serve as observed realizations of optimal messages ($O^*$) for our coverage-based evaluation. Table~\ref{tab:email_pipeline_all} summarizes the pre-processing pipeline.

\textbf{Human Personalization Strategies.}
To characterize the qualitative structure of expert personalization, we analyzed the strategies employed by SDRs across the Private Enterprise dataset (Fig.~\ref{fig:email_strategy}). The three most common strategies were: (i) \textbf{industry-based personalization}, tailoring content to sector-specific trends and pain points; (ii) \textbf{persona-based personalization}, adapting the value proposition to the recipient's organizational role; and (iii) \textbf{activity-based personalization}, leveraging behavioral signals such as webinar attendance or prior engagement. Appendix Fig.~\ref{fig:actual_emails} provides qualitative examples showing how SDRs adapt the same product positioning differently across recipients with distinct needs and pain points.

\section{Results \& Analysis}
\label{sec:results}
We evaluate six LLMs, \textbf{Claude Sonnet 4.6}~\citep{anthropic2026claudesonnet46}, \textbf{GPT-4o} and \textbf{GPT-4o-mini}~\citep{openai2024gpt4o}, \textbf{GPT-5.4} and \textbf{GPT-5.4-mini}~\citep{openai2026gpt54}, and \textbf{Qwen2.5-72B}~\citep{yang2024qwen25}, each with and without the web-search tool, and 2 deep-research agents, \textbf{STORM}~\citep{shao-etal-2024-assisting} and \textbf{ODR}~\citep{langchain2025opendeepresearch}, both built on Qwen2.5-72B, yielding 8 configurations with web search (\Cref{tab:ground-truth-artifacts}, \Cref{tab:enterprise-artifacts}) and a matching set of no-search ablations for the LLMs. Three datasets are used to evaluate these configurations: the publicly collected SDR-Bench success stories, evaluated on a stratified sample spanning four major industry types, \emph{Technology}, \emph{Manufacturing}, \emph{Energy}, and \emph{IT}, so that industry-level trends can be reported separately, supplemented with a random sample across the remaining industries to preserve the corpus's overall diversity;  enterprise sales emails from two cohorts, a \emph{Healthcare} company with under \$1B in revenue and a \emph{Technology} company exceeding \$10B in revenue, and transcripts of enterprise sales calls from the \emph{Technology} company. \Cref{tab:ground-truth-artifacts} reports coverage on the success stories and \Cref{tab:enterprise-artifacts} on the enterprise emails and sales calls. Average per-query token consumption and inference cost for every configuration are reported in Appendix~\ref{sec:appendix-results}. Please refer Fig.~\ref{fig:qualitative} for qualitative results.

\begin{table}[t]
\centering
\footnotesize
\setlength{\tabcolsep}{2.5pt}
\renewcommand{\arraystretch}{0.9}

\caption{\textbf{SDR-Bench.} WCS on success stories across four industries and on the whole set.}
\label{tab:ground-truth-artifacts}

\begin{tabular}{@{}lccccc@{}}
\toprule
Model & Tech. & Mfg. & Energy & IT & Agg. \\
\midrule
\multicolumn{6}{l}{\textit{Search}} \\
\textbf{Claude Sonnet 4.6} & \textbf{52.6} & \textbf{60.5} & \textbf{54.0} & \textbf{54.9} & \textbf{55.8} \\
GPT-5.4-mini              & 40.99 & 46.02 & 41.18 & 45.39 & 44.63 \\
GPT-5.4                   & 38.21 & 45.46 & 42.90 & 46.67 & 44.32 \\
STORM-QWEN-2.5            & 43.40 & 41.59 & 39.24 & 44.60 & 42.51 \\
GPT-4o-mini               & 33.30 & 38.54 & 37.07 & 38.05 & 37.46 \\
QWEN2.5-72B               & 32.09 & 36.75 & 37.04 & 38.21 & 36.84 \\
GPT-4o                    & 33.50 & 38.93 & 32.26 & 36.01 & 35.42 \\
ODR-QWEN-2.5              & 30.16 & 30.95 & 35.28 & 35.33 & 33.53 \\
\midrule
\multicolumn{6}{l}{\textit{Direct}} \\
Claude Sonnet 4.6         & 52.14 & 51.47 & 41.29 & 48.99 & 49.88 \\
GPT-5.4-mini              & 42.60 & 42.28 & 42.39 & 46.43 & 44.10 \\
GPT-5.4                   & 39.69 & 42.69 & 42.14 & 45.81 & 43.95 \\
GPT-4o                    & 35.52 & 38.15 & 36.62 & 37.44 & 38.15 \\
QWEN2.5-72B               & 31.28 & 33.09 & 36.28 & 38.19 & 35.67 \\
GPT-4o-mini               & 33.09 & 33.98 & 30.52 & 32.44 & 33.88 \\
\bottomrule
\end{tabular}
\end{table}

\begin{table}[t]
\centering
\footnotesize
\setlength{\tabcolsep}{3pt}
\renewcommand{\arraystretch}{0.88}

\caption{\textbf{Private Enterprise Dataset.} WCS on successful outreach emails and sales call transcripts.}
\label{tab:enterprise-artifacts}

\begin{tabular}{@{}lccc@{}}
\toprule
Model & \makecell{Health.\\($<$\$1B)} & \makecell{Tech.\\($>$\$10B)} & Calls \\
\midrule
\multicolumn{4}{l}{\textit{Search}} \\
\textbf{Claude Sonnet 4.6} & \textbf{60.89} & \textbf{61.33} & \textbf{42.85} \\
GPT-5.4-mini              & 43.67 & 52.66 & 35.72 \\
GPT-5.4                   & 48.71 & 53.02 & 33.61 \\
STORM-QWEN-2.5            & 32.27 & 39.24 & 33.80 \\
GPT-4o-mini               & 39.14 & 45.63 & 29.68 \\
QWEN2.5-72B               & 36.72 & 36.43 & 29.57 \\
GPT-4o                    & 40.62 & 45.17 & 26.19 \\
ODR-QWEN-2.5              & 30.41 & 39.82 & 26.92 \\
\midrule
\multicolumn{4}{l}{\textit{Direct}} \\
Claude Sonnet 4.6         & 42.93 & 54.26 & 39.71 \\
GPT-5.4-mini              & 42.01 & 54.09 & 34.53 \\
GPT-5.4                   & 46.53 & 55.01 & 36.03 \\
GPT-4o                    & 45.09 & 46.40 & 32.59 \\
QWEN2.5-72B               & 25.57 & 41.24 & 31.34 \\
GPT-4o-mini               & 30.15 & 42.95 & 30.05 \\
\bottomrule
\end{tabular}
\end{table}

\subsection{Discussion of Empirical Findings}

We observe several notable trends. First, \textbf{Claude Sonnet 4.6} leads all models with an aggregate WCS of 55.8\% on the SDR-Bench success stories, a 25.0\% relative improvement over the next best model, GPT-5.4-mini (44.6); yet even the best model recovers only about half the strategic content of the ground-truth success stories. We refer to this persistent gap as the \emph{personalization plateau}: the gap between the personalization capabilities of current agents and expert human senders. The plateau is broad rather than family-specific: the four GPT-family models cluster between 35.4 and 44.6 (overlapping bootstrap CIs for the weaker ones; Appendix~\ref{sec:confidence-intervals} Table~\ref{tab:bootstrap}), and the deep-research agents STORM (42.5) and ODR (33.5) do not escape it despite much higher cost.

The evaluation on \textbf{enterprise emails and sales calls} (\Cref{tab:enterprise-artifacts}) shows the same plateau: Claude leads on coverage for both cohorts: \emph{Healthcare} and \emph{Technology}, the GPT-5.4 family follows on the email data, and the deep-research agents and Qwen trail, yet even the strongest model recovers only about 60\% of the winning content on emails and well under half on transcripts. Absolute coverage is higher on the well-documented, larger Technology company than on the smaller, sparsely-documented Healthcare company for almost all the agent configurations (LLMs w/ and w/o web search $+$ DRAs). We next examine why: whether the plateau reflects a failure to find the right facts on the web, or a failure to reason with facts once found.

\subsection{The Bottleneck Is Retrieval, Not Reasoning}
\label{sec:retrieval-bottleneck}

\looseness=-1 The \textbf{LLM + Web Search} configuration is motivated by the premise that account-specific evidence (facts that make a pitch successful) must be retrieved rather than recalled from parametric memory. To test whether models actually convert retrieval into recovered strategic content, we re-ran every LLM with search disabled, holding prompts and inputs fixed (the \emph{Direct} rows of Table~\ref{tab:ground-truth-artifacts} and Table~\ref{tab:enterprise-artifacts}).

\looseness=-1 \textbf{Weaker models do not exploit search.} Removing web search changes WCS by less than one point for GPT-5.4 ($44.32 \rightarrow 43.95$), GPT-5.4-mini ($44.63 \rightarrow 44.10$), and QWEN-2.5-72B ($36.84 \rightarrow 35.67$), and even \emph{improves} GPT-4o ($35.42 \rightarrow 38.15$). If retrieval were supplying the deal-winning specifics,
removing it should sharply reduce coverage; instead WCS is nearly invariant
or rises, indicating these models fall back on prior knowledge or generic
firmographic content rather than converting search into strategic content.
\textbf{Claude is the exception.} It is the only model whose coverage
depends substantially on search: removing it costs Claude $17.96$ points on
Healthcare email outreach and $7.07$ on Tech, by far the largest such gaps of
any model, with a matching drop from $55.80$ to $49.88$ on the success stories.
Search also helps most where documentation is sparsest: on the well-documented Tech cohort, adding search is redundant or mildly harmful for weaker models, whereas on the sparser Healthcare cohort several weaker models gain substantially (Table~\ref{tab:enterprise-artifacts}). 

\textbf{Pre-training leakage is not driving WCS.}
A memorization probe, splitting SDR-Bench into pre- and post-2024 stories relative to
GPT-4o's training cutoff, shows negligible WCS differences across the split (STORM: $0.42$ vs.\ $0.43$; GPT-4o: $0.36$ vs.\ $0.36$), ruling out memorization of the specific ground-truth artifact as a confound.

Having shown that most models fail to \emph{find} the relevant content through open search, we next isolate whether the plateau is a retrieval failure or a reasoning failure: can models synthesize a winning pitch when handed the relevant facts directly? We asked expert SDRs for 50 questions they routinely ask when researching a deal (spanning customer, seller, and product), issued these as queries through the temporally constrained simulator, and compiled the results into four expert-curated research
reports: one per element (seller, product and customer), and one combining all three. We supply each report at generation time to the weakest model (Qwen2.5-72B) and the
strongest (Claude), each in Direct and +Search configurations, and report
$\Delta$WCS relative to the no-report baseline in Table~\ref{tab:expert-report}.

\looseness=-1 \textbf{Given the expert-curated content, models improve substantially}, and the gain persists even in the +Search configuration, where the model already has live web access: an expert-curated report still lifts WCS well beyond what the model's own search achieves on the same account (gains of $+5.3$ and $+3.3$ WCS for QWEN (Direct / $+$Search) and $+7.0$ and $+6.6$ for Claude (Table~\ref{tab:expert-report}). The bottleneck is therefore not the ability to reason and synthesize once content is available, but query formulation and information retrieval, knowing what to search for. Models can convert strategic content into stronger pitches once it is surfaced, whether by their own search or an expert's; they simply do not know how to surface it themselves.

\looseness=-1 \textbf{Seller and product evidence drive the gain; customer evidence
alone does not.} Reports about the seller and the product account for almost the entire improvement, whereas the customer-only report is flat-to-negative in three of four configurations (Table~\ref{tab:expert-report}), despite customer research being the part of the task that most resembles classical personalization. Facts about the customer in isolation appear to steer models toward restating the prospect's business rather than
constructing an argument for why the product addresses their needs; the gain materializes only once seller and product information, facts that do not vary across prospects, are also provided. Combining all three sources yields the most robust gain, indicating they are complementary once the model is anchored on what to look for.

\begin{table}[t]
\centering
\footnotesize
\setlength{\tabcolsep}{4pt}
\renewcommand{\arraystretch}{0.85}

\caption{\textbf{Expert-guided research reports.} $\Delta$WCS from expert-curated reports covering the customer, seller, product, or all three. Results are relative to each configuration's no-report baseline. All results reported as averages over 3 runs.}
\label{tab:expert-report}

\begin{tabular}{l cc cc}
\toprule
 & \multicolumn{2}{c}{QWEN-2.5-72B} & \multicolumn{2}{c}{Claude Sonnet 4.6} \\
\cmidrule(lr){2-3}\cmidrule(lr){4-5}
Report supplied & Direct & $+$Search & Direct & $+$Search \\
\midrule
$+$ Customer & $-2.5$ & $-3.6$ & $+0.9$ & $-2.4$ \\
$+$ Seller   & $+5.2$ & $+1.8$ & $+5.9$ & $+1.1$ \\
$+$ Product  & $\mathbf{+5.5}$ & $+2.8$ & $+5.7$ & $+1.7$ \\
$+$ All      & $+5.3$ & $\mathbf{+3.3}$ & $\mathbf{+7.0}$ & $\mathbf{+6.6}$ \\
\bottomrule
\vspace{-3mm}
\end{tabular}
\end{table}

\subsection{The Plateau Persists in a Discriminative Task}
\label{sec:discriminative}

Since WCS credits only content which matches the single recorded winner $V^*$, a low score could in principle reflect a competent model that produced a valid but different winning argument, one absent from the ground truth. The personalization plateau would then be an artifact of matching generated content to ground truth rather than a real capability gap. We test this directly with a discriminative task whose scoring does not depend on matching $V^*$ at all; the agent must recognize a winning email rather than reconstruct it.

\textbf{Setup.} We set up the discriminative task as an $N$-way forced choice. Each instance presents $N=5$ real cold outbound sales emails, one that received a positive response (a reply or booked call) and four that did not, all sent to prospects in the \emph{same industry}, and the agent must identify the one that succeeded out of the 5 emails. 

\begin{table}[t]
\centering
\caption{\textbf{Discriminative $N$-way accuracy} ($N=5$: one successful email vs.\ four same-industry unsuccessful cold outbounds), for the web-search LLMs and the deep-research agents, alongside their WCS on enterprise emails and success stories.}
\label{tab:discrimination}
\footnotesize
\setlength{\tabcolsep}{4pt}
\begin{tabular}{lccc}
\toprule
Model & \shortstack{$N$-way\\Acc.\ (\%)} & \shortstack{Email-Succ.\\WCS} & \shortstack{Story-Agg.\\WCS} \\
\midrule
Claude Sonnet 4.6 & 45.8 & 61.11 & 55.80 \\
GPT-5.4           & 40.5 & 50.87 & 44.32 \\
GPT-5.4-mini      & 39.0 & 48.17 & 44.63 \\
GPT-4o            & 37.0 & 42.90 & 35.42 \\
GPT-4o-mini       & 37.0 & 42.39 & 37.46 \\
STORM             & 36.2 & 35.76 & 42.51 \\
QWEN-2.5-72B      & 35.0 & 36.58 & 36.84 \\
ODR               & 32.2 & 35.12 & 33.53 \\
\midrule
\multicolumn{4}{l}{Spearman $\rho$ (Acc.\ vs.\ WCS): $0.97$ (Email), $0.84$ (Story)} \\
\bottomrule
\end{tabular}%
\end{table}

\looseness=-1 \textbf{Even when choosing the successful email, the best model is right less than half the time.} Models struggle on this task despite having to only \emph{recognize} the winning email, not author it, and random guessing already yields $20\%$. The best frontier model, Claude Sonnet 4.6, picks the winning email out of five only $45.8\%$ of the time (Table~\ref{tab:discrimination}), less than half. The GPT-5.4 and GPT-4o families occupy the middle ($37$--$40.5\%$), and QWEN-2.5-72B sits at $35.0\%$; the deep-research agents fare worst, with STORM at $36.2\%$ and ODR at $32.2\%$ despite their far higher cost.

These results show that the models are not able to reliably differentiate a successful email from unsuccessful emails, reinforcing the finding of a personalization plateau. The discriminative task shows that the personalization plateau is a genuine limitation in personalization capability and not a byproduct of measuring coverage against a recorded winner. The same ranking of models also emerges here as on the generative task, discrimination accuracy and WCS are rank-correlated at Spearman $\rho = 0.97$ ($p < 0.001$) against enterprise-email WCS and $\rho = 0.84$ ($p \approx 0.01$) against SDR-Bench story WCS, showing that the two tasks agree on which models are weaker, but both place every model far short of reliable personalization.

\subsection{Token Usage, Cost, and Performance}
\label{sec:cost-analysis}

\begin{figure}[htbp]
    \centering
    \includegraphics[width=\linewidth]{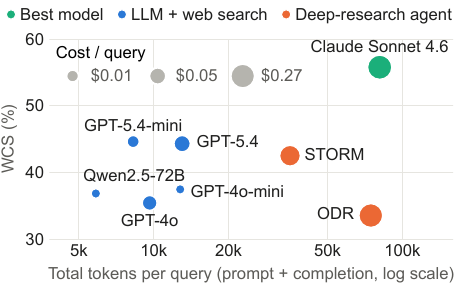}
    \caption{Token usage vs.\ performance on SDR-Bench. Each point is a
    configuration; the $x$-axis is total tokens per query (log scale), the $y$-axis is WCS, and bubble area encodes inference cost per query.}
    \label{fig:tokens}
\end{figure}

\looseness=-1 Per-query token usage, inference cost, and WCS for every configuration are reported in Table~\ref{tab:cost-analysis} and visualized in Figure~\ref{fig:tokens}. GPT-5.4-mini is the most efficient frontier model ($44.6$ WCS at $\sim$8k tokens), whereas the deep-research agents spend one to two
orders of magnitude more tokens than LLM+search baselines without escaping the personalization plateau. Claude Sonnet~4.6 achieves the highest WCS ($55.8$) at a token budget comparable to STORM and ODR, both of which trail it by more than 13
points.

\subsection{Human Alignment and Validation}

To check that WCS reflects real-world quality, we validate the Coverage Judge against independent human raters and test its rankings against the judgement of professional sales representatives.

\looseness=-1 We conducted a human study on 20 success stories, scoring 100 model responses from five web-search agents: STORM, ODR, Claude Sonnet~4.6, and the best-performing GPT and Qwen configurations. Three independent human annotators, blinded to model identity, scored coverage following the exact protocol of our LLM-based Coverage Judge.

The study yields three convergent signals on Judge fidelity. First, the
Judge tracks human scores with strong rank correlation (Spearman's
$\rho = 0.7435$, $p < 0.0001$), holding across models (Claude: $0.7685$;
ODR: $0.7768$; GPT: $0.7575$; Qwen: $0.7330$; STORM: $0.6963$). Second, the
Judge \emph{preserves model ordering}: both human- and Judge-graded WCS
rank Claude $>$ STORM $\approx$ GPT $>$ Qwen $>$ ODR, so absolute-score
differences do not distort comparative conclusions. Third, the Judge is
systematically more conservative than human raters (Claude: $48.06$ vs.\
$55.29$; GPT: $37.94$ vs.\ $44.04$; ODR: $31.10$ vs.\ $41.69$), ruling out
score inflation and establishing WCS as a conservative, empirically
validated proxy that tracks human judgement.

\looseness=-1 We show that the WCS-based ranking transfers to expert SDR judgment through
two field studies with 12 senior SDRs from the partner enterprises. The
first study measures \emph{per-pitch usefulness}: whether a generated pitch point
would be used verbatim in real outreach. The second measures
\emph{strategy-level overlap} between model output and SDR-authored gold
standards, probing per-point quality and overall strategic match
respectively.

\textbf{Per-pitch usefulness.} 12 SDRs used our best-performing model
(Claude Sonnet 4.6) with SDR-Arena to generate pitch points for $200+$ new
prospects inside their normal outbound pipeline, auditing output on
accounts they were actively working. Each generated pitch point was rated on whether
it (a) reflected genuine understanding of the prospect's pain points and (b)
was usable without rewriting; \textbf{48.2\%} met both criteria, roughly half of agent output being expert-grade in live deployment which is consistent
with the personalization plateau identified in results.

\textbf{Gold-standard alignment.} Senior SDRs ($\geq 10$ years of
experience, $\geq 5$ at the firm) from five enterprises authored reference ``gold-standard'' strategies for 30 products across 5 prospects each, without seeing any
model output. Correlating expert-overlap scores against automated WCS
across models yields Pearson $r = 0.816$, showing the WCS ranking transfers
to senior-SDR judgment without re-tuning the rubric per enterprise.

Together, these studies establish that our framework's outputs are both factually grounded and meaningful in live sales contexts.
\label{sec:human}

\section{Conclusion}

\looseness=-1 We introduced SDR-Arena, a framework for benchmarking two-party generative personalization at scale, and SDR-Bench, a public corpus of $\sim$50k customer success stories, to measure whether an LLM can recover the strategic content that historically induced a receiver to act. Across frontier LLMs and deep-research agents, we find a consistent personalization plateau: even the strongest model recovers only about half the strategic content of outreach known to have succeeded. An independent discriminative task and field deployment with professional sales representatives confirm this reflects a real shortfall rather than an artifact of scoring against a single ground truth. The gap traces to retrieval, not reasoning: models improve substantially when handed the facts an expert researcher would find, but cannot surface those facts through open web search on their own. We release SDR-Arena and SDR-Bench to support reproducible research on this
capability.

\paragraph{Limitations:} Our study has several limitations. Our evidence comes from English-language sales outreach across a large but still a limited set of enterprises, so the personalization plateau we observe may not transfer to other languages, or receivers outside the enterprise sales context we study. Second, ground-truth pitch points and their coverage are produced by LLM-based extraction and judging; although both are validated against human annotators (extraction precision 0.92 / recall 0.97; judge Spearman $\rho=0.74$), residual model bias may remain. Finally, the WCS measures argument-content coverage grounded in the ELM and deliberately does not capture stylistic delivery or multi-turn conversational quality.

\subsection{Ethical Considerations}\label{sec:broader-impacts}
Our benchmark could help build more manipulative outreach, but we believe openly measuring such systems is safer than leaving them unevaluated. The proprietary emails were processed under access control with all PII anonymized or redacted, and are \emph{not} released. SDR-Bench contains only published, enterprise-level success stories, and our web search tool honors robots.txt. An Acceptable Use Policy will prohibit using our artifacts for spam, undisclosed AI-written correspondence, or phishing.

\FloatBarrier
\raggedbottom 
\bibliographystyle{acl_natbib}
\bibliography{example_paper}

\appendix

\section{Replication Details}
\label{sec:appendix-replication}

\subsection{SDR-Bench Dataset Curation}
\label{sec:sdr-bench-details}

SDR-Bench is built from publicly available customer success stories. Starting
from $\sim$12{,}000 global enterprises with revenue exceeding \$1B, we resolved
company domains, located sitemaps, and harvested candidate success-story URLs
using path heuristics (e.g., \texttt{/customer-stories}, \texttt{/case-study}).
A multi-stage pipeline then removed non-text formats, generic or listicle pages,
articles without a verifiable publication date or identifiable product, and
anonymized stories whose customer was not explicitly named. Each stage and its
surviving count are reported in Table~\ref{tab:sdr_bench_filter}, yielding a final
corpus of 50{,}000 stories across 22 industries.

\begin{table}[ht]
\centering
\footnotesize
\caption{Filtration criteria and surviving counts for the public data collected to build SDR-Bench.}
\label{tab:sdr_bench_filter}
\begin{tabular}{@{}p{0.66\linewidth} r@{}}
\toprule
\textbf{Filtration Criteria} & \textbf{Count} \\
\midrule
Domains for companies with $>$\$1B revenue & $\sim$30k \\
B2B companies with $>$\$1B revenue & 12{,}080 \\
Companies whose URLs were discovered & 3{,}500 \\
Total webpages found & $\sim$7M \\
Candidate success story URLs & $\sim$400k \\
After Qwen content filter (listicle/parent/\newline generic/no-date pages removed) & $\sim$60k \\
After removing non-specific customers & 50,000 \\
\bottomrule
\end{tabular}
\end{table}

\subsection{Enterprise Sales Email Processing}
\label{sec:email_pers_strategies}

\begin{table}[h]
\centering
\caption{Processing of Enterprise Sales Email Data}
\label{tab:email_pipeline_all}
\footnotesize
\begin{tabularx}{\linewidth}{>{\raggedright\arraybackslash}X cc}
\toprule
\textbf{Metric / Artifact Category} & \textbf{Healthcare} & \textbf{Tech} \\ \midrule
Number of SDRs & 3 & 124 \\
Total Emails Collected & 48,150 & 609,191 \\
Deduplication & 31,034 & 186,379\\
Sales Outreach Emails & 24,506 & 90,809 \\
Successful Outreach (Call Scheduled) & 354 & 13,236 \\ \bottomrule
\end{tabularx}
\end{table}
\subsubsection{Filtering and Extraction}

Let $\mathcal{E} = \{e_1, e_2, \ldots, e_N\}$ denote the raw corpus of sales
emails. We apply a three-stage filtering pipeline:
\begin{itemize}[leftmargin=*, topsep=2pt, itemsep=1pt]
    \item \textbf{Language filtering:} remove non-English emails via language
    detection, yielding $\mathcal{E}_{\text{en}} \subset \mathcal{E}$.
    \item \textbf{Deduplication:} remove duplicate templates using exact matching
    and fuzzy string comparison, yielding
    $\mathcal{E}_{\text{dedup}} \subset \mathcal{E}_{\text{en}}$.
    \item \textbf{Intent classification:} an LLM judge
    $\mathcal{J}: e \rightarrow \{0,1\}$ retains valid outreach
    ($\mathcal{J}(e)=1$) and discards generic, administrative, or non-sales
    email, giving
    $\mathcal{E}_{\text{filtered}} = \{e \in \mathcal{E}_{\text{dedup}} :
    \mathcal{J}(e)=1\}$.
\end{itemize}

For each $e \in \mathcal{E}_{\text{filtered}}$ we use an LLM to extract the
strategies employed, $\text{Strat}(e) \subseteq \mathcal{S}$, and the fine-grained
pitch points $\mathbf{pp}(e) = \{pp_1, \ldots, pp_k\}$; the latter constitute the
ground truth against which agent outputs are scored. Each annotated sample stores
the target company $T_i$, sender company $S_i$, email body $E_i$, timestamp $t$,
product $P$, strategy labels $\text{Strat}(e)$, and pitch points
$\mathbf{pp}(e)$. The resulting strategy-frequency distribution $P(s)$ and the
product-conditional distribution $P(s \mid \text{Product}_k)$ summarize how human
SDRs operationalize personalization in practice.

\subsubsection{Personalization Strategy Taxonomy}

To characterize personalization strategies we used a two-step pipeline:
(i) domain experts manually annotated a seed set of emails to identify recurring
patterns, and (ii) an LLM extracted and clustered strategies from 500 randomly
sampled emails, which were reconciled with the expert annotations into a unified
taxonomy. A strategy $s \in \mathcal{S}$ denotes the primary information source
used to establish relevance between the seller's value proposition and the
buyer's needs. The strategy space $\mathcal{S} = \{s_1, \ldots, s_{10}\}$
comprises:

\begin{itemize}[leftmargin=*, topsep=2pt, itemsep=1pt]
    \item \textbf{Industry-based:} industry-specific trends, pain points,
    competitors, or case studies.
    \item \textbf{Event-based:} trigger events (funding, M\&A, launches, earnings,
    news) signaling timely needs.
    \item \textbf{Technology-based:} the recipient's current tech stack, for
    replacement, integration, or complementary solutions.
    \item \textbf{Lead-activity-based:} direct actions by the lead (downloads,
    webinar attendance, pricing-page visits, demos).
    \item \textbf{Buying-group-activity-based:} collective actions by the lead's
    team or buying committee.
    \item \textbf{Geography-based:} location or regional regulatory context
    (e.g., GDPR, CCPA).
    \item \textbf{Lead-persona-based:} the lead's role, title, or responsibilities
    mapped to role-specific pain points.
    \item \textbf{Firmographics-based:} company-level metrics (headcount growth,
    revenue, department size).
    \item \textbf{Relationship-based:} existing customer relationships, cross-sell
    or upsell.
    \item \textbf{None:} generic outreach lacking recipient-specific context.
\end{itemize}

\begin{figure}[!t]
    \centering
    \includegraphics[width=\linewidth]{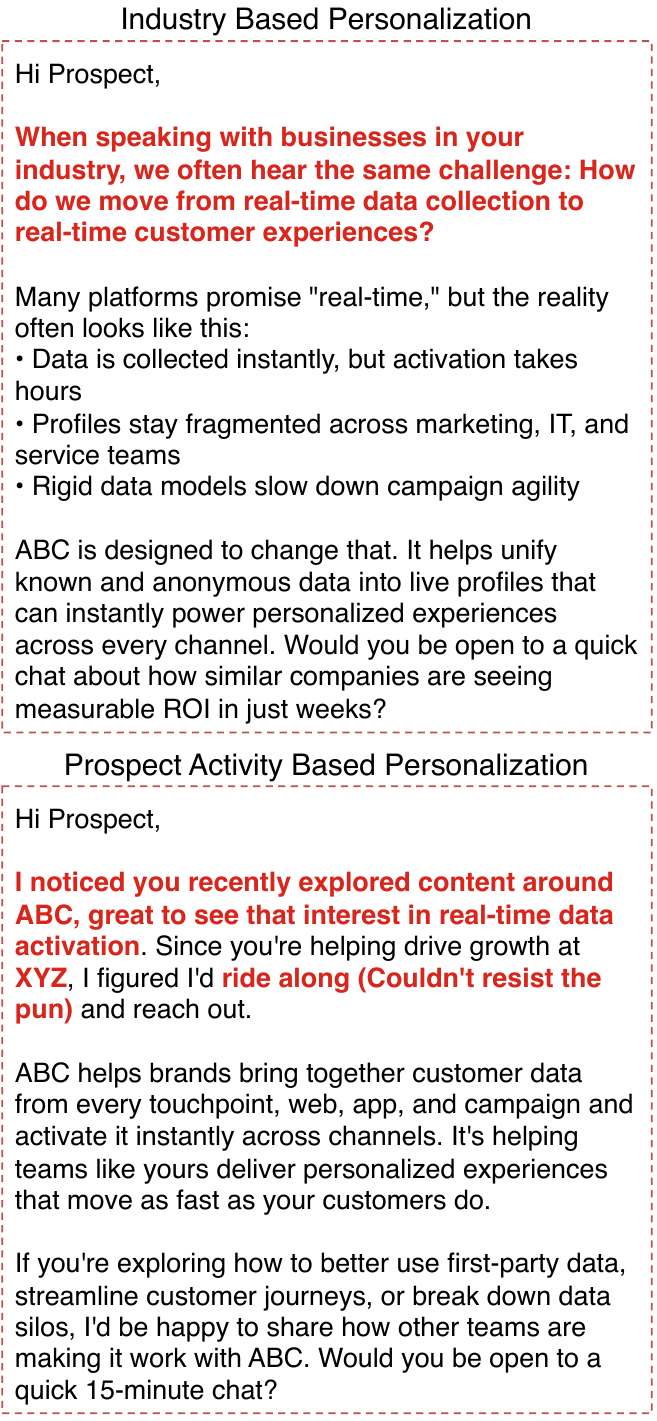}
    \caption{Personalization in actual sales emails: SDRs adapt the same product
    positioning across recipients with distinct needs and pain points.}
    \label{fig:actual_emails}
\end{figure}

\subsection{Pitch-Point Extraction and Human Validation}
\label{sec:human-study-pitch-points}

Ground-truth pitch points $V^*$ are extracted from each artifact with GPT-4o under
an ontological prompt (App.~\ref{sec:prompt-library}): each pitch point must follow
the triad \emph{Product/Service $\rightarrow$ Specific Pain Point $\rightarrow$
Value Proposition/Mechanism} and be validated with an exact evidence quote from the
source text. To assess extraction quality, three annotators reviewed the extracted
pitch points on a random sample of 30 customer success stories, flagging
hallucinated points (false positives) and missing points (false negatives).
Table~\ref{tab:pitch_extraction_counts} reports the resulting counts: precision
$0.92$, recall $0.97$, and F1 $0.95$, indicating strong alignment with expert
judgment.

\begin{table}[ht]
\centering
\footnotesize
\caption{Human evaluation of LLM-extracted pitch points on 30 customer success
stories, assessing precision (factual consistency, hallucination flagging) and
recall (missed pitch points).}
\label{tab:pitch_extraction_counts}
\begin{tabular}{@{}cccccc@{}}
\toprule
\textbf{TP} & \textbf{FP} & \textbf{FN} & \textbf{Precision} & \textbf{Recall}
& \textbf{F1} \\
\midrule
138 & 11 & 3 & 0.92 & 0.97 & 0.95 \\
\bottomrule
\end{tabular}
\end{table}

\subsection{Prompt Library}
\label{sec:prompt-library}

We provide the exact prompts used for LLM-based judging
(Box~1), pitch-point generation (Box~2), and
ground-truth pitch-point extraction (Box~3). Placeholders in
\texttt{\{braces\}} or \texttt{<{}<{}<angle brackets>{}>{}>} are substituted at runtime.

\begin{tcolorbox}[
    colback=boxBodyBg, colframe=boxTitleBg, coltitle=white,
    fonttitle=\bfseries, title=Box 1: LLM-as-a-Judge Evaluation Prompt,
    label=box:judge, sharp corners, boxrule=1pt, enhanced,
    attach boxed title to top left={yshift=-2mm, xshift=2mm},
    boxed title style={colback=boxTitleBg, sharp corners}, breakable
]
\footnotesize
\textbf{SYSTEM --- Role:} You are a Senior Sales Enablement Evaluator. Your goal is
to determine if an AI agent has successfully extracted the ``Winning Pitch Points''
from a customer success story.

\vspace{0.4em}
\textbf{INPUT DATA}
\begin{enumerate}[leftmargin=*, noitemsep, topsep=0pt]
    \item \textbf{Ground-Truth (GT) Winning Points} --- the proven, specific facts
    that won the deal: \texttt{<{}<{}<gt\_pitch\_points\_str>{}>{}>}
    \item \textbf{Candidate Pitch (Predicted)} --- the points generated by the AI
    agent: \texttt{<{}<{}<candidate\_pitch\_points\_str>{}>{}>}
\end{enumerate}

\textbf{TASK:} For each Ground-Truth point, determine how well the Candidate Pitch
covers it. Grade on \textbf{Sales Effectiveness} and \textbf{Factual Precision}.

\vspace{0.4em}
\textbf{SCORING RUBRIC (0--5):}
\begin{itemize}[leftmargin=*, noitemsep, topsep=0pt]
    \item \textbf{0 (Miss / Irrelevant):} completely misses this concept.
    \item \textbf{1 (Marketing Fluff):} vaguely mentions the topic (e.g.,
    ``improved efficiency'') with no specific substance.
    \item \textbf{2 (Topic Match):} correct Product or Pain Point, but misses the
    specific Solution or Outcome.
    \item \textbf{3 (Implied / Soft Match):} captures the core value proposition,
    but misses ``hero evidence'' (specific numbers, names, mechanisms).
    \item \textbf{4 (Strong Sales Argument):} captures core value \emph{and} the key
    mechanism/outcome; may miss a minor detail.
    \item \textbf{5 (Strategic Bullseye):} perfect extraction capturing
    Product + Pain + Value + Specific Metric/Evidence, essentially verbatim.
\end{itemize}

\vspace{0.4em}
\textbf{OUTPUT FORMAT (JSON only):}
\begin{lstlisting}[style=promptbox]
{
  "evaluations": [
    {
      "gt_point_id": <int>,
      "gt_summary": "<short_summary_of_gt_point>",
      "best_match_candidate_text": "<text_from_candidate_or_null>",
      "score": <0-5>,
      "reasoning": "<concise_sales_analysis>"
    }
  ]
}
\end{lstlisting}
\end{tcolorbox}

\begin{tcolorbox}[
    colback=boxBodyBg, colframe=boxTitleBg, coltitle=white,
    fonttitle=\bfseries, title=Box 2: Pitch-Point Generation Prompt,
    label=box:generation, sharp corners, boxrule=1pt, enhanced,
    attach boxed title to top left={yshift=-2mm, xshift=2mm},
    boxed title style={colback=boxTitleBg, sharp corners}, breakable
]
\footnotesize
I am a BDR at \texttt{\{seller\}} (\texttt{\{seller\_website\}}). I want to sell
\texttt{\{seller\}} products to \texttt{\{customer\}}
(\texttt{\{customer\_website\}}). How should I pitch
\texttt{\{products\}} to \texttt{\{customer\}}? Generate 3 targeted pitch points
and value propositions that address the customer's pain points, which I can send
to \texttt{\{customer\}} or their CEO / leads / decision makers. Provide reasoning
for each pitch point.

\vspace{0.4em}
\textbf{Requirements for each pitch point:}
\begin{itemize}[leftmargin=*, noitemsep, topsep=0pt]
    \item Specific to the target company (use real facts you discover).
    \item Connects the sender's products/services to the target's needs/pain points.
    \item Atomic --- one clear value proposition per point.
    \item Compelling and actionable.
\end{itemize}

After researching, generate 3 pitch points, each with a single value proposition
addressing the customer's pain point. Each must be unique (the same product or pain
point may recur, but the value proposition must differ).

\vspace{0.4em}
\textbf{Respond with a JSON array of pitch-point strings:}
\begin{lstlisting}[style=promptbox]
[
    "First pitch point text here",
    "Second pitch point text here",
    "Third pitch point text here"
]
\end{lstlisting}
Strictly adhere to the JSON format above.
\end{tcolorbox}

\begin{tcolorbox}[
    colback=boxBodyBg, colframe=boxTitleBg, coltitle=white,
    fonttitle=\bfseries, title=Box 3: Pitch-Point Extraction Prompt,
    label=box:extraction, sharp corners, boxrule=1pt, enhanced,
    attach boxed title to top left={yshift=-2mm, xshift=2mm},
    boxed title style={colback=boxTitleBg, sharp corners}, breakable
]
\footnotesize
You are an expert sales analyst. Analyze the following success-story content for
\texttt{\{company\_name\}} (the seller) and their product
\texttt{\{product\_name\}}. Story content: \texttt{\{story\_html\}}.

\vspace{0.4em}
Extract specific, causal pitch points that explain \emph{why} the product was
successful, and provide exact quotes from the text as evidence.

\vspace{0.4em}
\textbf{STRICT FORMAT:} return a JSON object with a single key
\texttt{"pitch\_points"} containing a list of objects of the form:
\begin{lstlisting}[style=promptbox]
{
    "summary": "<Product/Service> solved <Specific Pain Point> of <Customer>
     because of <Value Proposition/Mechanism>[, resulting in <Quantitative Result>]",
    "evidence": [
        "Exact quote from the text supporting the pain point...",
        "Another exact quote..."
    ]
}
\end{lstlisting}

\textbf{GUIDELINES:}
\begin{enumerate}[leftmargin=*, noitemsep, topsep=0pt]
    \item \textbf{Product/Service:} use the specific product name if available.
    \item \textbf{Specific Pain Point:} name the exact problem; distinguish
    ``slow/manual work'' from ``inaccuracy/errors'' --- these are distinct.
    \item \textbf{Value Proposition/Mechanism:} the specific feature that solved it.
    \item \textbf{Quantitative Result:} if the text gives a number (e.g.,
    ``70\% faster''), you must include it in the summary.
    \item \textbf{Evidence:} quote the original text; do not paraphrase.
\end{enumerate}

\vspace{0.4em}
\textbf{Example output:}
\begin{lstlisting}[style=promptbox]
{
    "pitch_points": [
        {
            "summary": "The Aldec G3 solved the high energy costs of T.Z. Osborne
             because of its PowerTubes technology that recovers kinetic energy,
             reducing consumption by 20%.",
            "evidence": ["Facing rising energy costs...",
             "The PowerTubes technology recovers kinetic energy,
              reducing consumption by 20%"]
        },
        {
            "summary": "Oracle Integration solved the issue of manual entry errors
             at Careem because of its automated GRN matching feature.",
            "evidence": ["Manual entry led to mismatches and compliance gaps.",
             "Automating GRN matching has significantly reduced errors."]
        }
    ]
}
\end{lstlisting}
\end{tcolorbox}

\FloatBarrier
\section{Supplementary Results and Analysis}
\label{sec:appendix-results}

\begin{table}[H]
\centering
\footnotesize
\setlength{\tabcolsep}{2pt}
\renewcommand{\arraystretch}{1.0}
\caption{Per-query token usage, inference cost, and WCS on the SDR-Bench
evaluation set. \textbf{Search} = temporally-constrained web search;
\textbf{Direct} = no tool use. Deep-research agents are search-only. Token counts
are mean prompt/completion tokens per query; costs use public API list prices at
evaluation time. The deep-research pipelines (STORM, ODR) consume one to two orders
of magnitude more tokens than the LLM${+}$search baselines for comparable or lower
WCS.}
\label{tab:cost-analysis}
\begin{tabular}{@{}ll rrr r@{}}
\toprule
\textbf{Model} & \textbf{Mode} & \textbf{Prompt} & \textbf{Compl.}
& \textbf{Cost} & \textbf{WCS} \\
 & & \textbf{Tok.} & \textbf{Tok.} & \textbf{(\$)} & \\
\midrule
STORM & Search & 29k & 6.3k & 0.135 & 42.51 \\
ODR   & Search & 66k & 8.6k & 0.250 & 33.53 \\
\midrule
\multirow{2}{*}{\textbf{Claude Sonnet 4.6}}
      & Search & 79k   & 2.2k  & 0.270  & \textbf{55.80} \\
      & Direct & 0.3k  & 1.2k  & 0.019  & 49.88 \\
\midrule
\multirow{2}{*}{GPT-5.4-mini}
      & Search & 7.6k  & 0.7k  & 0.009  & 44.63 \\
      & Direct & 0.3k  & 0.67k & 0.003  & 44.10 \\
\midrule
\multirow{2}{*}{GPT-5.4}
      & Search & 12.1k & 0.9k  & 0.044  & 44.32 \\
      & Direct & 0.3k  & 1.15k & 0.018  & 43.95 \\
\midrule
\multirow{2}{*}{GPT-4o-mini}
      & Search & 12.2k & 0.6k  & 0.002  & 37.46 \\
      & Direct & 0.3k  & 0.4k  & 0.0003 & 33.88 \\
\midrule
\multirow{2}{*}{Qwen2.5-72B}
      & Search & 5.6k  & 0.25k & 0.002  & 36.84 \\
      & Direct & 0.3k  & 0.35k & 0.0003 & 35.67 \\
\midrule
\multirow{2}{*}{GPT-4o}
      & Search & 9.2k  & 0.4k  & 0.027  & 35.42 \\
      & Direct & 0.3k  & 0.55k & 0.006  & 38.15 \\
\bottomrule
\end{tabular}%
\end{table}

\subsection{Qualitative Example}
\label{sec:qualitative-example}

\begin{figure*}[t]
    \centering
    \includegraphics[width=0.98\textwidth]{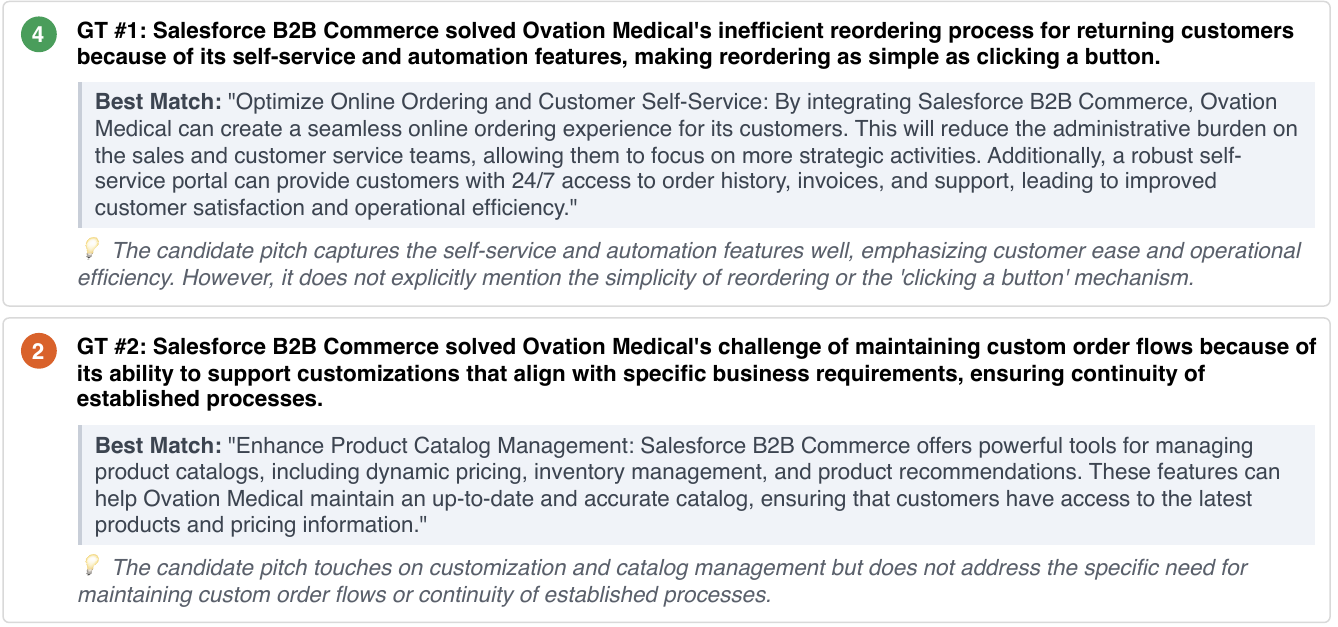}
    \caption{\textbf{Qualitative example} of the Coverage Judge: ground-truth pitch
    points matched against the agent's generated pitch points and scored on the
    $0$--$5$ rubric of Appendix~\ref{sec:prompt-library}, illustrating how partial
    topic matches are separated from strategic bullseyes.}
    \label{fig:qualitative}
\end{figure*}

\subsection{Bootstrap Confidence Intervals}
\label{sec:confidence-intervals}

\begin{table}[H]
\centering
\footnotesize
\caption{Bootstrap estimates of WCS on the SDR-Bench evaluation set
($1{,}000$ resamples), with per-configuration means and 95\% confidence intervals.
The weaker GPT-family and Qwen intervals overlap (the personalization plateau),
while Claude Sonnet~4.6 separates clearly above the rest.}
\label{tab:bootstrap}
\begin{tabular}{@{}lcc@{}}
\toprule
\textbf{Model} & \textbf{Mean WCS} & \textbf{95\% CI} \\
\midrule
Claude Sonnet 4.6 & 0.5580 & [0.5238, 0.5906] \\
GPT-5.4-mini      & 0.4488 & [0.4265, 0.4700] \\
GPT-5.4           & 0.4457 & [0.4216, 0.4683] \\
STORM             & 0.4251 & [0.4003, 0.4517] \\
GPT-4o-mini       & 0.3767 & [0.3567, 0.3977] \\
Qwen-2.5-72B      & 0.3684 & [0.3478, 0.3878] \\
GPT-4o            & 0.3562 & [0.3345, 0.3775] \\
ODR               & 0.3353 & [0.3139, 0.3570] \\
\bottomrule
\end{tabular}
\end{table}

\subsection{Success-Story Distribution by Industry}
\label{sec:story-distribution}

\begin{figure}[H]
    \centering
    \includegraphics[width=\linewidth]{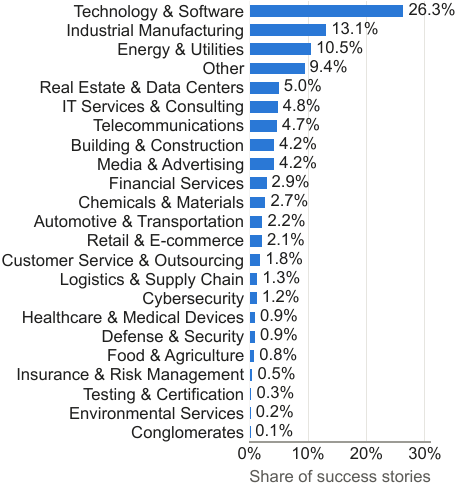}
    \caption{SDR-Bench: Distribution of the count of success stories across different industries, complementing the company-level distribution in
    Figure~\ref{fig:sdrbench_company_dist_by_industry}. The corpus spans a broad
    range of sectors, with technology, manufacturing, and energy the most
    represented.}
    \label{fig:sdrbench_story_dist_by_industry}
\end{figure}

\end{document}